# ISNA-Set: A novel English Corpus of Iran NEWS

Mohammad Kamel, Hadi Sadoghi-Yazdi
Center of Excellence on Soft Computing and Intelligent Information Processing
Ferdowsi University of Mashhad
Mashhad, Iran
*{mohammad.kamel, h-sadoghi}@mail.um.ac.ir*

***Abstract*— News agencies publish news on their websites all over the world. Moreover, creating novel corpuses is necessary to bring natural processing to new domains. Textual processing of online news is challenging in terms of the strategy of collecting data, the complex structure of news websites, and selecting or designing suitable algorithms for processing these types of data. Despite the previous works which focus on creating corpuses for Iran news in Persian, in this paper, we introduce a new corpus for English news of a national news agency. ISNA-Set is a new dataset of English news of Iranian Students News Agency (ISNA), as one of the most famous news agencies in Iran. We statistically analyze the data and the sentiments of news, and also extract entities and part-of-speech tagging.**

## I. INTRODUCTION

There are a number of news agencies all over the world that broadcast and publish local and international news daily. Today's people understand that even the events and news of other countries may have direct effect on their lives; therefore, they continuously track the wide range of news sources. Traditionally, an important source of information were newspapers that had effective editorial stage in their publishing. However, social networks and media are the main sources of information today. *Citizen journalism* is a new concept which emphasizes that even a normal citizen can be a source of information for news. Although the social networks such as Twitter and Facebook can be considered as a source of news nowadays, most of people prefer to rely on valid news agencies as daily news resources.

There are a number of news agencies in Iran and Iranian people check them regularly to get the most recent national and international news. Furthermore, many smaller newspapers also employ them as one of their main resources. Iranian Students News Agency (ISNA), Islamic Republic News Agency (IRNA), and Islamic Republic of Iran Broadcasting (IRIB) are among the most famous news agencies in Iran. These agencies publish daily news in different categories. These outputs become a valuable resource for a vast range of processing over the time. For instance, determine the amount of influence of a special type of event on people for finding the causality relations between events. However, the textual data itself can be an input corpus for Natural Language Processing (NLP) research projects. The text and body of a news is written by journalists who are fluent in that language and have high level of writing skills. Furthermore, usually an editor revises the text from the language and content points of view; hence, the text of news are notably clean for processing and do not need any preprocessing stages.

Creating a novel corpus is a traditional trend in the field of NLP. For instance, [1] [2] [3] are among the most well-known works in this area. However, most of the efforts on creating corpus for Iran news focus on Persian language [4] [5]. DOTIR [6] and BIJAN KHAN [7] are some examples of Persian corpuses. Creating a new corpus is challenging in many ways. First, the strategy of collecting data is important since downloading all news of a news agency relies heavily on the features of the corresponding hosting server. After that, parsing the files is a challenge due to the complex structure of news websites. Finally, selecting or designing suitable algorithms for processing of data is significant. Entity extraction, part-of-speech tagging, sentiment analysis, and word modeling are among the most famous NLP algorithms.

The sentiment analysis in NLP is the task of finding the mood and emotion of a writer based on his/her textual data. A number of semantic oriented and machine learning oriented [8] [9] methods are introduced for this task. The traditional method of extracting the sentiment of a text is creating a list of positive and negative words and calculate the polarity by these lists. Traditionally, researchers have supervised viewpoints to this task [9], but there is a tendency to employ unsupervised and semi-supervised methods recently [10]. Online news are expected to be media professionals and therefore, neutral from the sentiment point of view. However, the personal biases of journalists, editors and also the news can add the sentiment to the news content.

Extracting the sentiment of news illustrates highly valuable information about the events over a period of time, the viewpoint of a media or news agency to these events and the existing bias between journalists and editors. This problem counts as an interesting task in the field of NLP. For instance, multi-label classification of news headlines is considered in the task of "Affective Text Analysis" in *SemEval-2007* [11]. SWAT [12] is among the solutions that have impressive results on this task. This supervised lexicon-based method creates a word-emotion mapping and classifies the whole text by this dictionary. As a relevant work, the idea of employing emotion-term is proposed in [13]. However, these word-level sentiment analysis techniques are not suitable for all topics since one word can have different meanings in different contexts. The emotion-topic model [14] is proposed to overcome this problem. Latent topic models, e.g. latent Dirichlet allocation, is employed to consider the different meanings of a word in this model. As specific-purpose solutions [15] and [16] use several multi-label topic models for sentiment analysis. Recently, a number of effective sentiment analysis methods are proposed both for general-purpose [17] [18] and domain-specific problems [19] [20].

One of the main NLP tasks is document classification. For this purpose, we need to extract appropriate features from each document and employ them as the input of a classifier. There are several effective machine learning techniques for data classification for example Support Vector Machines (SVM). However, feature extraction is a challenging part since each document can be processed in different levels of abstraction and from syntax and semantic points of view. Word2Vec is an effective method to map each word to a fixed-size numeric vector. Combining the corresponding vectors of main vocabularies of a document is a suitable feature of that document.

We introduce a new corpus is this paper. ISNA-Set is a new dataset of all English news of ISNA as one of the most famous news agencies in Iran. There are 1015 news in the corpus which are divided into six categories. We numerically analyze the data and release the dataset consisting of the histogram of entities, sentiment analysis, part-of-speech and different attributes of news. All of the mentioned materials are publically released and could be downloaded from here.

We describe the techniques of collecting the data and corpus generation is Section 2. Next, we statistically analyze the ISNA-Set and its categories in Section 3.

## II. Data Creation of ISNA-Set Corpus

In this section, we introduce the method of data gathering, data cleaning, data processing and model generation. Fig. 1 illustrates the process of data generation. In the first step, we create a list of URLs by analyzing the archive of the English section of ISNA website. After that, a python-based web crawler is employed to download all raw HTML files. Next, an HTML parser is employed for dividing the raw HTML files into the main sections of the news pages to create consistent JSON files. The next step is data cleaning, which consists of converting dates to timestamps (e.g. converting *"Sat / 24 June 2017 / 14:18"* to *"1495668060"*) and normalizing the textual data, such as lemmatization and stemming. Lastly, a Word2Vec model is generated based on the clean text to map the words of the dictionary to a fixed-length real-value space.

We use *wget* to download all raw HTML pages of the English version of ISNA. These raw HTML files are parsed by *BeautifulSoup* which is an effective *Python* Library for parsing any structural data. The structure of the data is changed in this phase and the unwanted data such as menus, headers, and footers are ignored and the related sections, e.g. news body, date and category are stored as JSON files. Table I. illustrates a sample of the contents of JSON files.

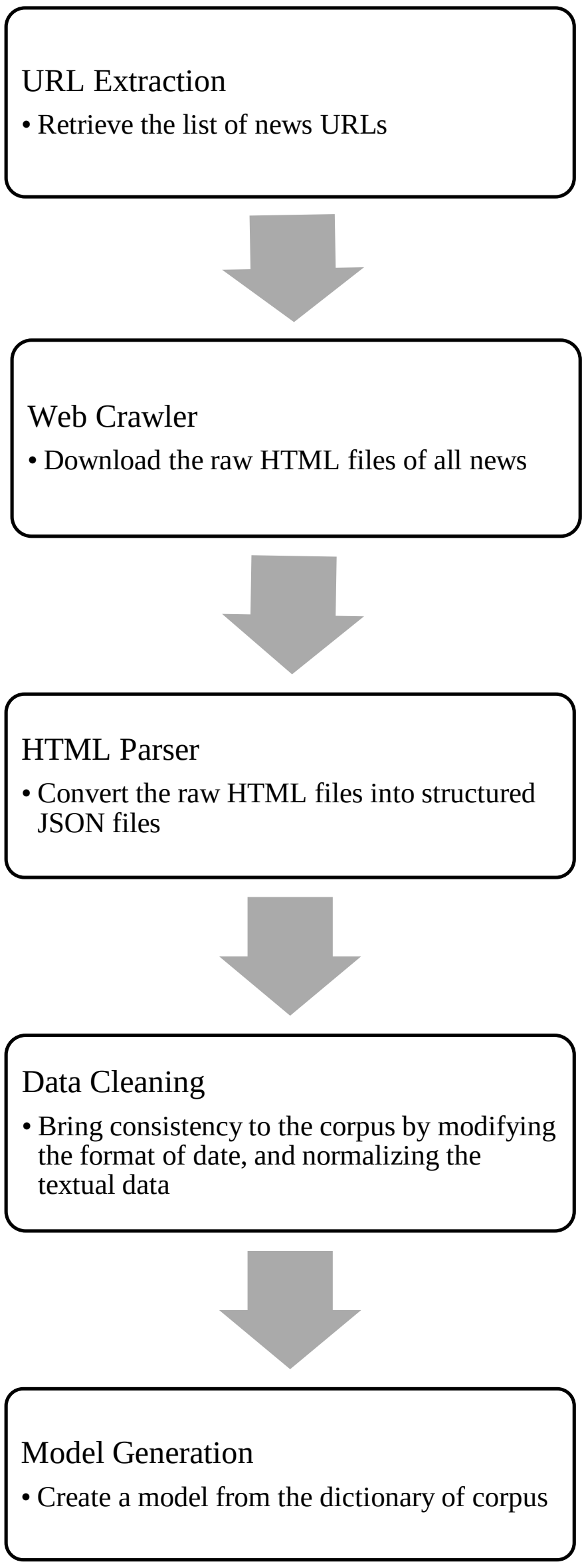

Fig 1 - The main steps in creating the ISNA-Set corpus

In the Data Cleaning phase, we need to prepare the data for further natural language and machine learning processing. By stemming and lemmatization, the size of the dictionary is reduced and processes become more effective. Furthermore, several non-ASCII characters exist in the news body that should be removed. We also need to convert the data from a human-friendly format to timestamp which is more appropriate for computations.

## III. Statistical Analysis

We analyze the statistics and main properties of ISNA-Set Corpus in this section. There are 1015 news in ISNA-Set. As Fig 2 shows, most of news have less than 400 words and the news with more than 1200 words are rare.

TABLE I. A SAMPLE CONTENT OF THE FINAL JSON FILE

```
{
"body": "Tehran (ISNA) –Iran's Economic Development Committee chairman …",
"code": "96030302144",
"sentiment": {
"body_s": -3,
        "title_s": 0
},
"tags": "Imad Khamis, Irans economic development committee, Saeed Ohadi, Syrian PM",
"entity": {
        "ORDINAL": "0",
…
},
"pos_tagging": {
        "ADV": "1",
        …
},
"date": "24 May 2017",
"category": "Politics",
"datetimestamp": 1495668060,
"title": "Iranian officials meet Syrian PM",
"url": "http://en.isna.ir/news/96030302144/",
"journalist": "71477"
}
```

Fig 3 illustrates the number of news in each day of the week. As the figure demonstrates, since Thursdays and Fridays are weekends in Iran, there is not any considerable number of news in these two days. Also, highest number of news belong to Saturdays and Mondays, as the first week days of Iran and the world.

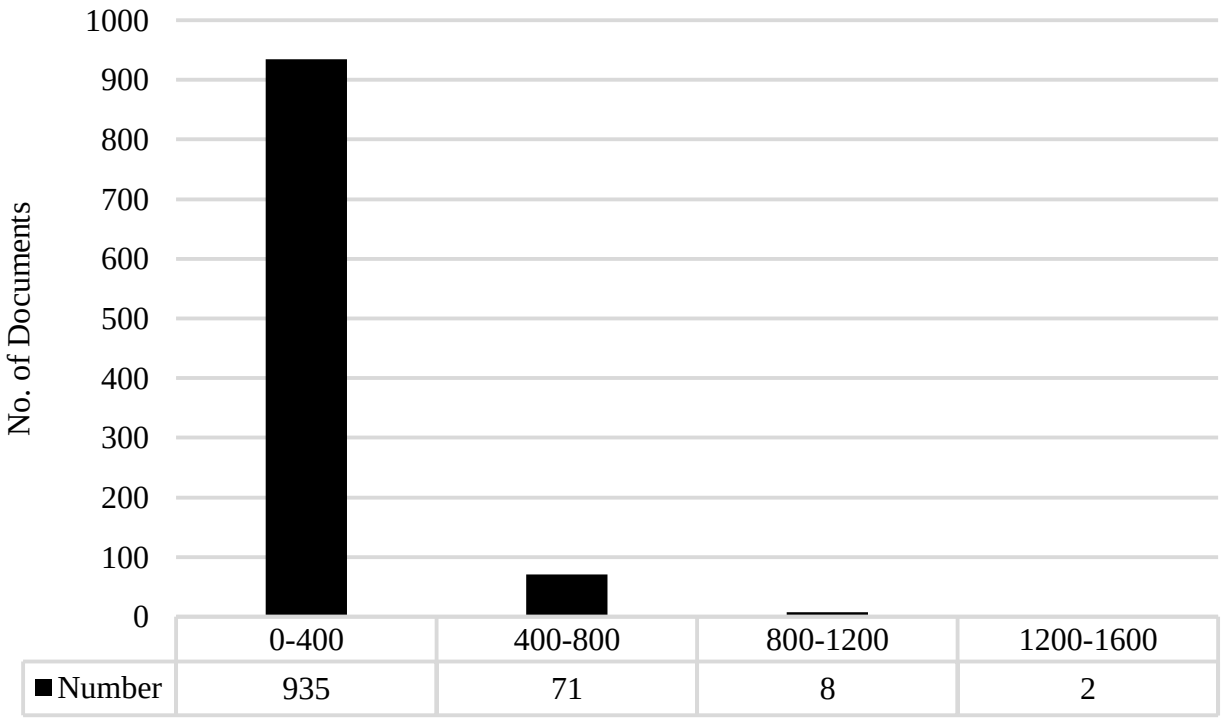

| | 0-400 | 400-800 | 800-1200 | 1200-1600 |
|---|---|---|---|---|
| Number | 935 | 71 | 8 | 2 |

Fig 2. The histogram of number of documents based on the number of words

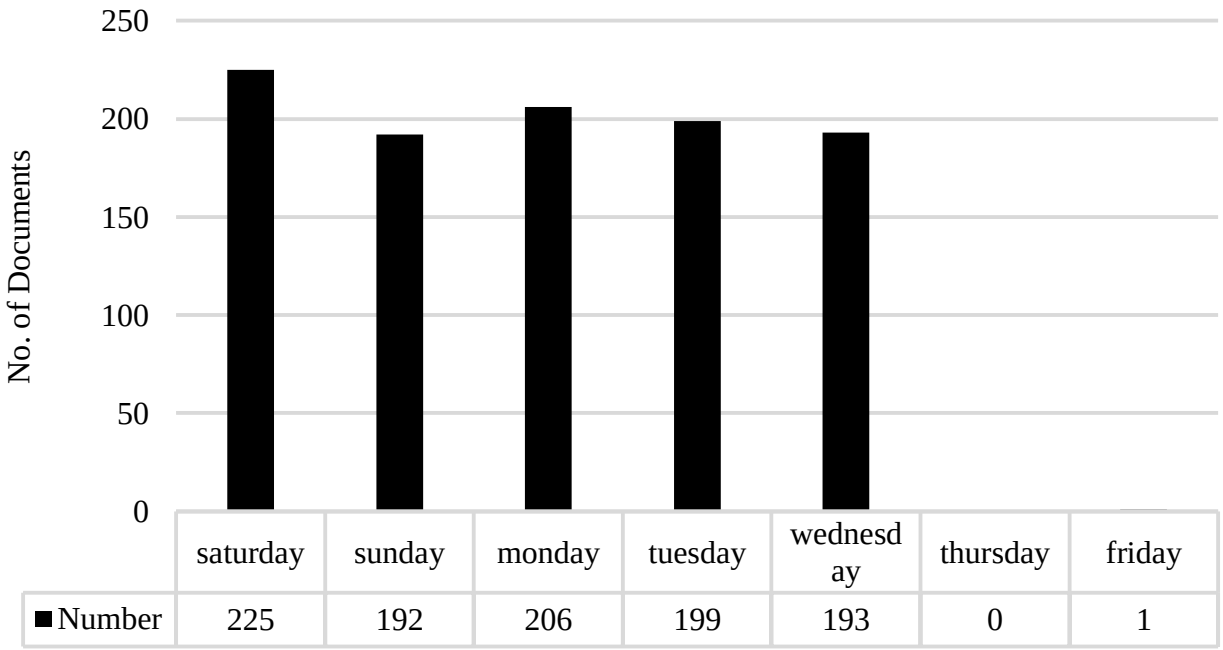

| | saturday | sunday | monday | tuesday | wednesday | thursday | friday |
|---|---|---|---|---|---|---|---|
| Number | 225 | 192 | 206 | 199 | 193 | 0 | 1 |

Fig 3. The number of new of ISNA-Set in each day of the week

Fig 4 shows the number of news in each day of the month. Although no particular pattern is found in this figure, we can conclude that the first and last days of months have less news in comparison with the other days.

After that, it is good to analyze the number of news in each month of the year. By considering Fig 5, the first and last months of year do not include considerable news in the corpus. Months eight and ten are the months of approval of budget in Iran and therefore, there are many economic and political news in these two months. Also month six stands in the third place since it is the last month of summer followed by fall in which the schools start. As the Fig 5 demonstrates, as we move forward from the first of summer to the end of it (month four to month six), the number of news increase. On the contrary, as we move forward from the start of winter to the end of it (month ten to month twelve), the number of news decrease significantly. The last month of winter is also the last month of year.

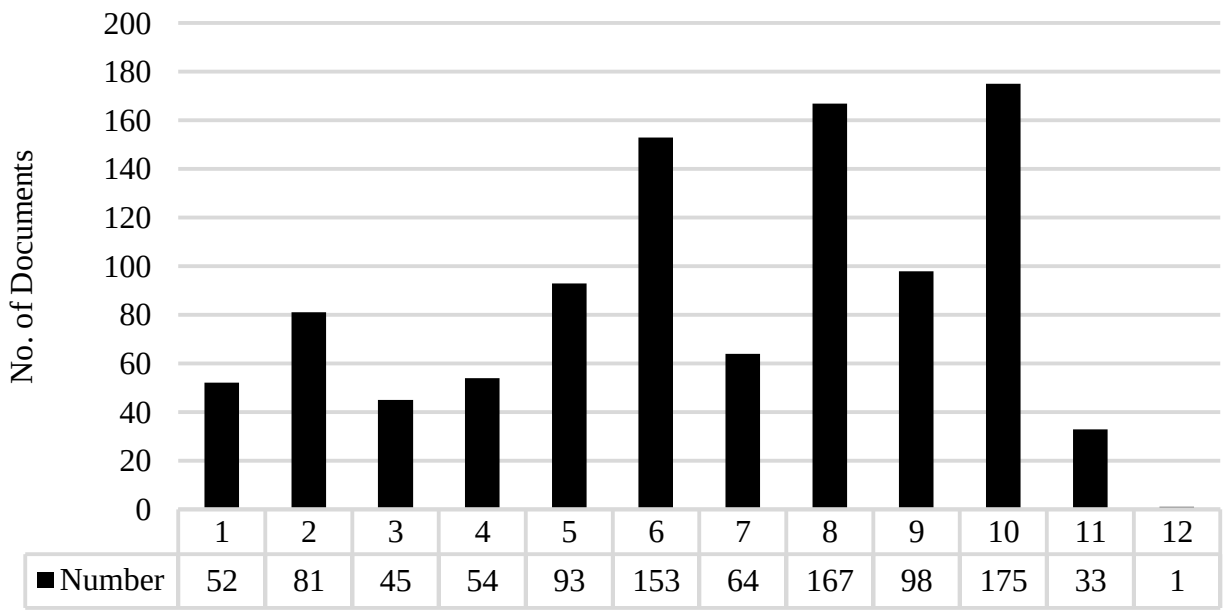

| | 1 | 2 | 3 | 4 | 5 | 6 | 7 | 8 | 9 | 10 | 11 | 12 |
|---|---|---|---|---|---|---|---|---|---|---|---|---|
| Number | 52 | 81 | 45 | 54 | 93 | 153 | 64 | 167 | 98 | 175 | 33 | 1 |

Fig 5. The number of new of ISNA-Set in each month of year

After that, we focus on different news categories. According to Fig 6, there are 1015 news in the ISNA-Set that most of them lie first in the *Politics,* and second, with a considerable difference from the first place, in *Economy* categories. Also, *Social* and *Sport* news categories are in minority.

Next, we need to analyze the length of each document. According to Fig 7, the majority of documents have less than *2K* characters and a document which have more than *5K* characters is rarely found in the corpus. The biggest set is the group of *415* news which contains *500-1000* tokens. For the number of characters more than 1k, as this number increases, the number of documents decreases most of the time.

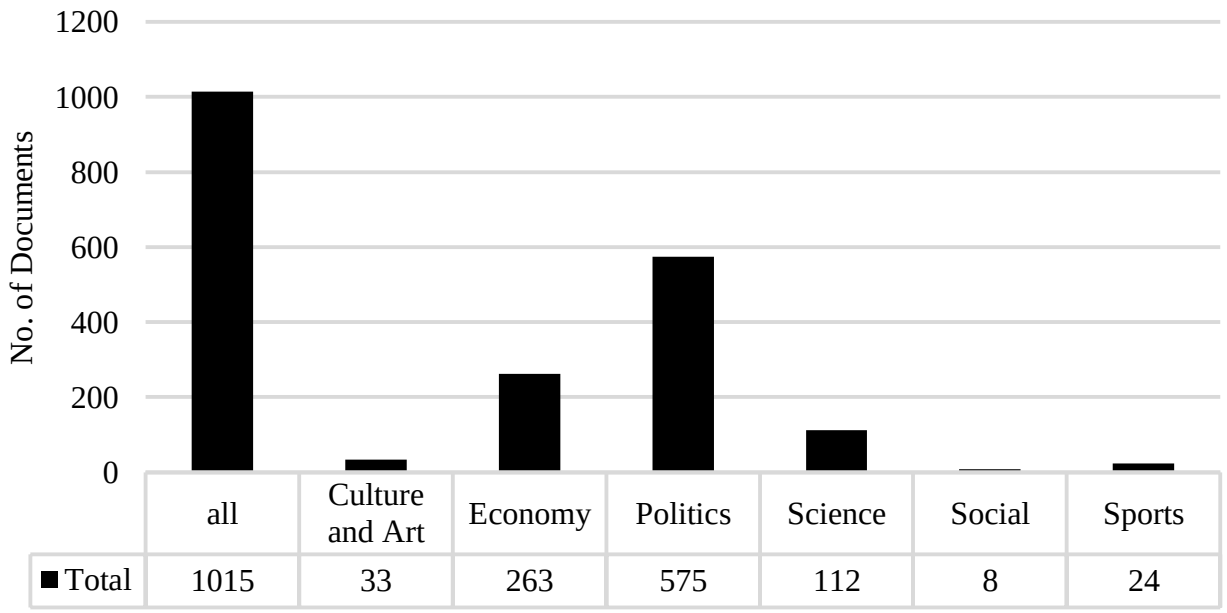

| | all | Culture and Art | Economy | Politics | Science | Social | Sports |
|---|---|---|---|---|---|---|---|
| Total | 1015 | 33 | 263 | 575 | 112 | 8 | 24 |

Fig 6. The histogram of ISNA-Set based on the number of characters

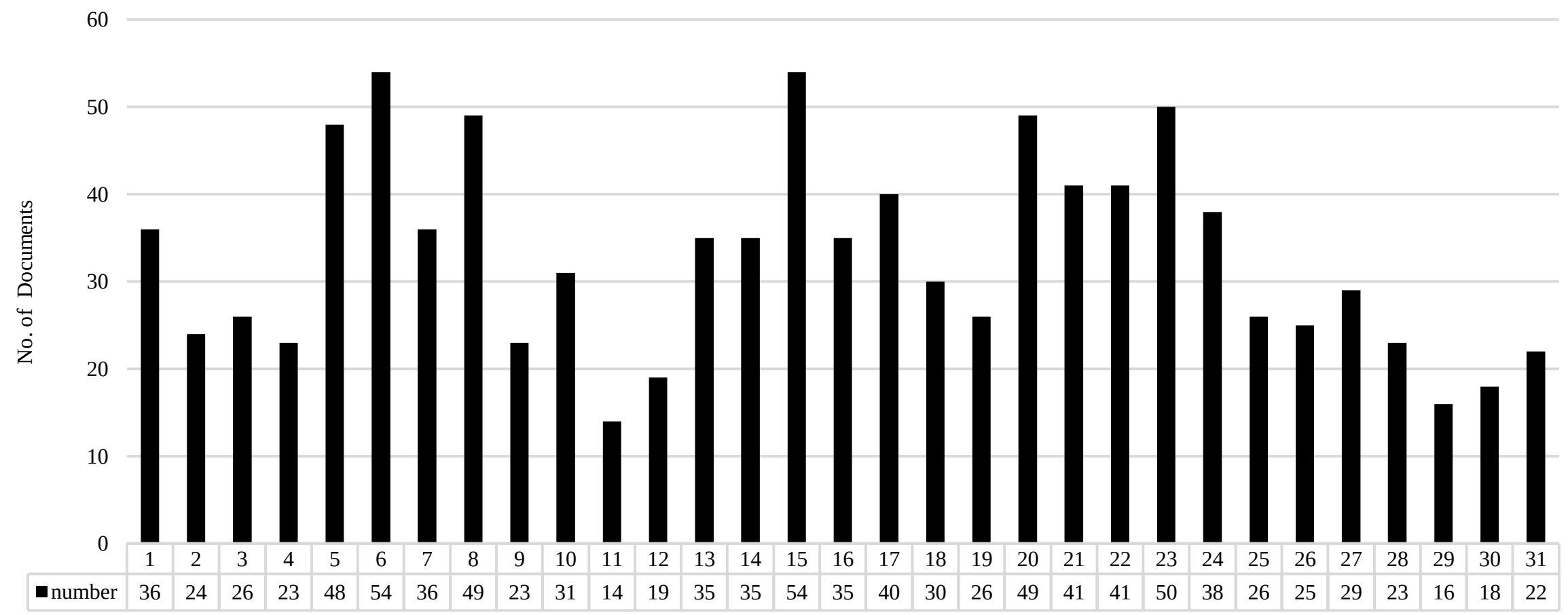

Fig 4. The number of new of ISNA-Set in each day of month

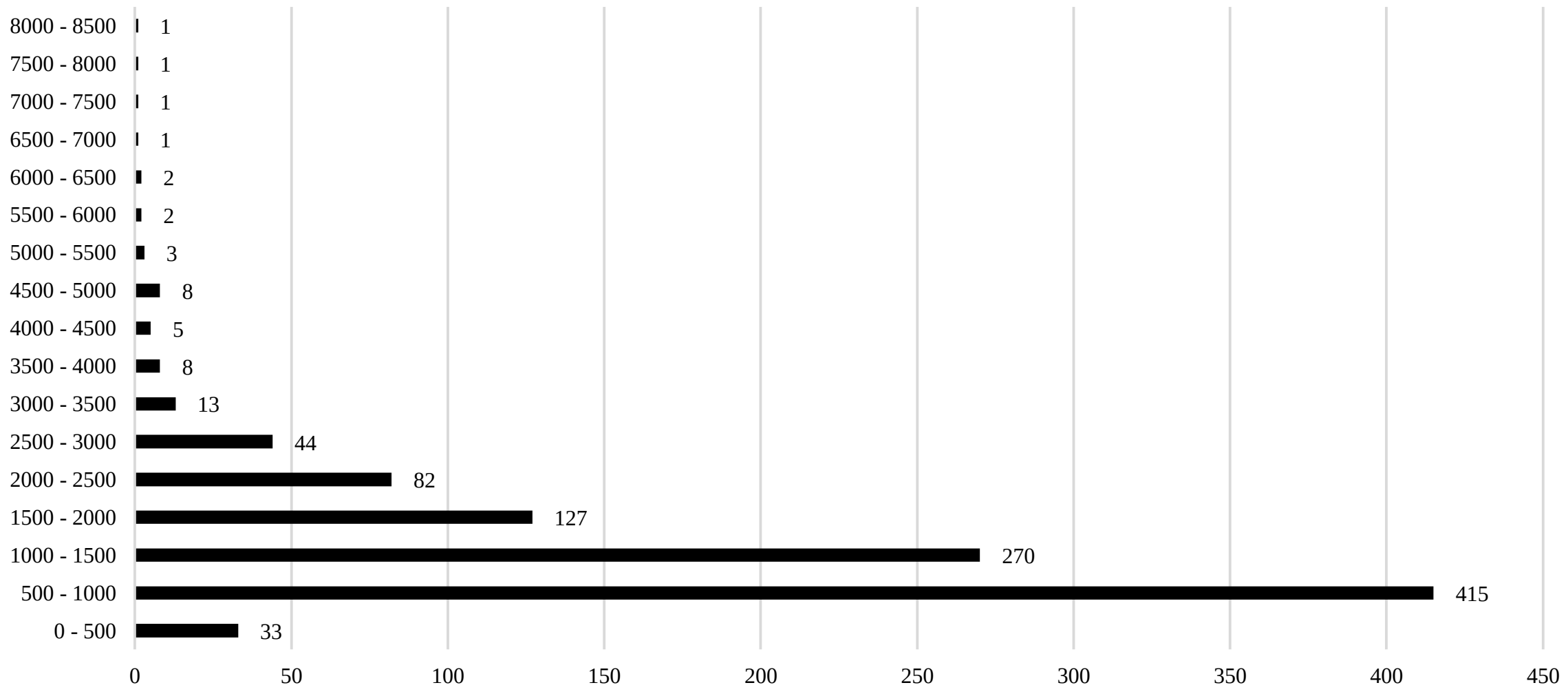

Fig 7. The histogram of ISNA-Set based on the number of characters

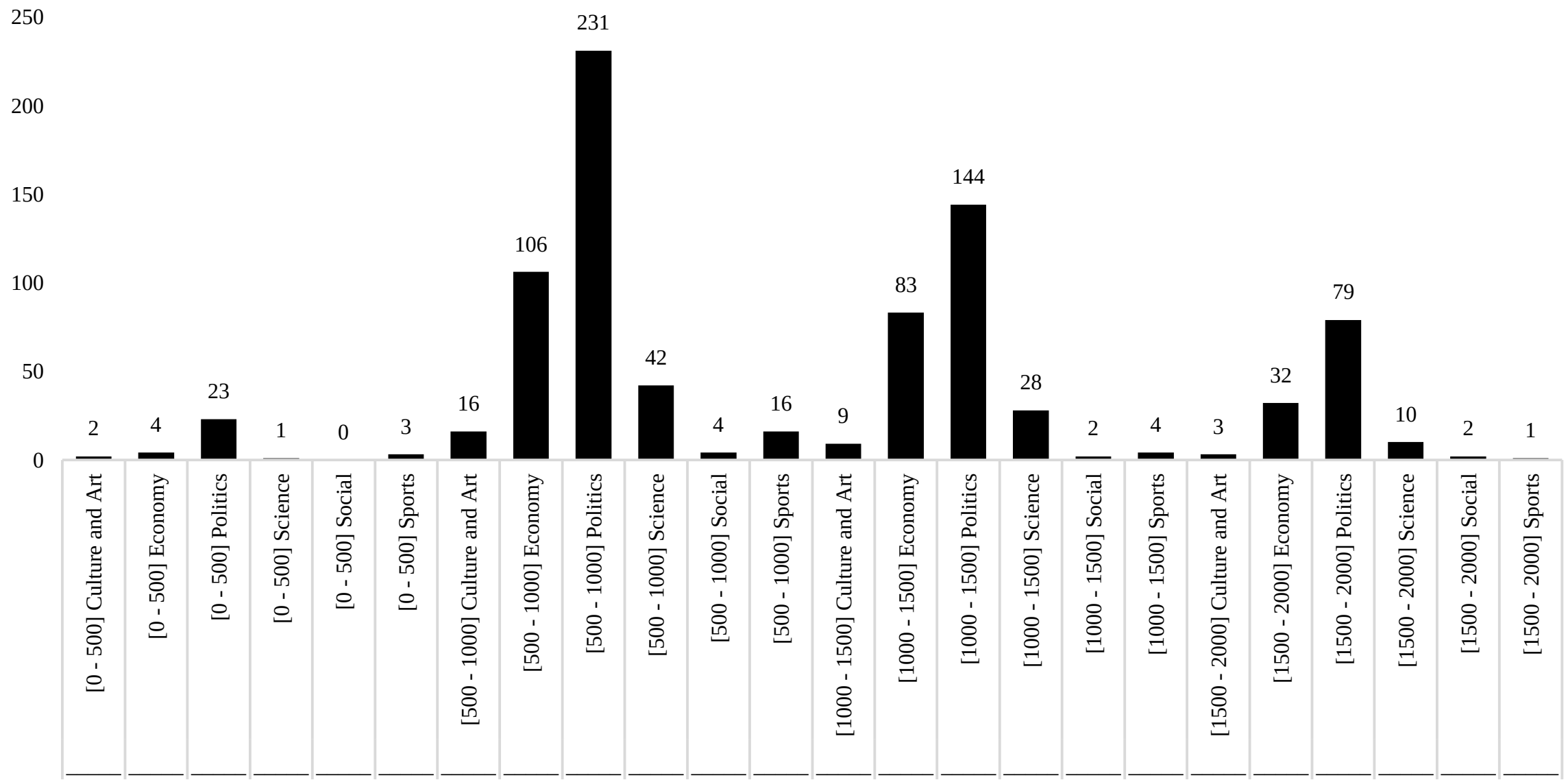

Fig 8. The histogram of ISNA-Set based on the number of words

Next, we analyze the number of characters per category. As the Fig 8 shows, Political news with number of characters between 500-1000 are in majority.

## Conclusion

We introduce a new corpus for English news of Iranian Students News Agency (ISNA) in this paper. All attributes of news such as body, header, date, journalist, and tags are extracted from the English version of ISNA website and are stored as JSON files. After that, further processing is performed to extract the sentiment of news, entity types, and part-of-speech. All of the mentioned materials are publically released and could be downloaded from here.

## Reerences